\documentclass[a4paper,10pt]{article}
\usepackage[preprint]{nips_2018}
\usepackage{amsmath}
\usepackage{amssymb}   
\usepackage{graphicx}  
\usepackage{hyperref}
\usepackage{subcaption}
\usepackage{booktabs}  
\usepackage{amsmath}
\usepackage{algorithm}
\usepackage[noend]{algpseudocode}
\usepackage{natbib}
\setcitestyle{numbers,open={[},close={]},citesep={,}}

\usepackage{placeins}

\DeclareMathAlphabet{\mathcal}{OMS}{cmsy}{m}{n}

\newcommand{\bigO}[1]{\ensuremath{\mathop{}\mathopen{}\mathcal{O}\mathopen{}\left(#1\right)}}  
\newcommand*\diff{\mathop{}\!\mathrm{d}}   

\newcommand{\subfig}[4]{
    \begin{subfigure}[h]{#1\textwidth}
    \centering 
    \includegraphics[width=\textwidth]{#2}
    \caption{#3}
    \label{#4}
    \end{subfigure}
}

\title{Spatiotemporal Prediction of Ambulance Demand using Gaussian Process Regression}

\author{Seth Nabarro\textsuperscript{a,b,*}
\qquad Tristan Fletcher\textsuperscript{a,b}
\qquad John Shawe-Taylor\textsuperscript{b}\\
\textsuperscript{a}InformedActions\\
\textsuperscript{b}University College London\\
*\texttt{ucabnab@ucl.ac.uk}}

\date{Received: date / Accepted: date}

\begin{document}
\maketitle

\begin{abstract}
Accurately predicting when and where ambulance call-outs occur can reduce response times and ensure the patient receives urgent care sooner. Here we present a novel method for ambulance demand prediction using Gaussian process regression (GPR) in time and geographic space. The method exhibits superior accuracy to MEDIC, a method which has been used in industry. The use of GPR has additional benefits such as the quantification of uncertainty with each prediction, the choice of kernel functions to encode prior knowledge and the ability to capture spatial correlation. Measures to increase the utility of GPR in the current context, with large training sets and a Poisson-distributed output, are outlined.
\end{abstract}


\section{Introduction}
Fast ambulance responses are often vital in achieving a good clinical outcome following a medical emergency. Indeed, ambulance response time has been shown to correlate with mortality \cite{O'Keeffe,Pell,Blackwell}. Accurate prediction of the density of the calls in time and space can improve readiness and therefore reduce response times. In addition, longer-term prediction of demand can inform staffing levels, ensuring that the workforce available is always sufficient to meet the demand.
Here we assert that the locations and times of previous emergencies will inform the anticipation of future call-outs. We seek to fit patterns of the demand in time and space, which can be extrapolated to generate predictions for future timesteps.
This paper begins with an overview of GPR, its suitability to spatiotemporal applications and how we go beyond standard GPR to improve performance with a challenging problem. We review existing work before describing how we evaluate predictive performance. Next, we outline our method for call-out prediction - spatiotemporal GP (STGP) - and present results, which are compared against those of the MEDIC benchmark. We end with some suggestions as to how this work might be exploited or extended.

\subsection{Gaussian Process Regression}
By definition, a Gaussian process is a stochastic process in which any finite subset of the random variables are Gaussian distributed \cite{MacKay}. They are commonly used as prior distributions in Bayesian inference. Covariance functions, which capture the similarity in output for a given pair of inputs, can be chosen based on known characteristics of the system, seasonality and trends for example.
GPR has exhibited promising performance in a wide variety of spatiotemporal applications. A non-exhaustive list includes: determining the distribution of available protein (in space and time) during post-transcriptional regulation \cite{Niu}; mapping brain structure developmental profiles of patients suffering from neurodegenerative diseases \cite{Hyun};  monitoring the spread of gastroenteritis \cite{Diggle}, influenza \cite{Senanayake} and malaria \cite{Mubangizi}.
A number of properties of GPR are thought to produce superior performance in spatiotemporal applications. Being a member of the Bayesian nonparametric class of methods, the output can be modelled without knowledge of all its dependencies or an assumed parametric form. Though, prior knowledge or intuition for types of spatial or temporal variation can be incorporated in expressive covariance functions. Further, large training sets are often required to capture spatiotemporal interactions and while traditional GPR is not scalable, recent research into faster alternatives makes GPR with large training sets accessible.
In addition to benefits of GPR listed, we see the ability to forecast with uncertainty as a significant advantage in the current application. Articulation of uncertainty allows any action made in light of predicted demand to be properly qualified.

\subsection{Challenges}
There are a number of challenges in applying GPR here:
\begin{enumerate}
\item {\bf Non-Gaussianity.} Standard GPR assumes a Gaussian distributed noise model. In predicting ambulance demand, the output is the number of counts in a given bin, and therefore a Poisson noise model is more appropriate.
\item {\bf Number of Training Instances.} As ambulance call-outs must be binnned in time and space, the number of training samples, $n$, scales cubically with the three-dimensional resolution. Any resolution sufficient to capture the spatiotemporal dynamics results in a large number of training samples. Standard GPR involves matrix inversion, with undesirable $\bigO{n^3}$ computational complexity.
\item {\bf Sparsity.} As observed by \cite{Zhou2}, the number of call-outs in most spatial bins is zero for most time steps. This can create a poor signal to noise ratio which causes difficulties during model training.
\end{enumerate} 

\subsection{GPR for Poisson Regression}
\label{sec:LGCP}
To make use of the desirable properties of GPR while correctly capturing the discrete nature of the output, we use a log-Gaussian Cox process (LGCP) \cite{Moller}. A Cox process is an inhomogeneous  Poisson process where the underlying rate $\lambda=\lambda(\mathbf{x})$ is a realisation of a stochastic process, denoted as $\rho(\mathbf{x})$. A LGCP extends a Cox process to model $\rho(\mathbf{x})$ as a transformed GP. The complete model description is
\begin{align}
\mathbf{f} &\sim \mathcal{GP}(\mu(\mathbf{X}), k(.,.))\\
\rho(\mathbf{x}) &= \exp\big(f(\mathbf{x})\big)\\
y_i | f(\mathbf{x}_i) &\sim \mathrm{Poisson}\big(\rho(\mathbf{x}_i)\big)
\end{align}
where $\mu(.)$ and $k(.,.)$ are the mean and covariance functions of the GP, $\mathbf{y}=\{y_i\}_{i=1}^n$ are the observed outputs, $\mathbf{f}=\{f_i\}_{i=1}^n$ is the GP model fit, and $\mathbf{X}$ is the design matrix. Note that the exponential transformation of $f(.)$ ensures non-negativity and thus a well-defined Poisson process.

\section{Scalable GPR}
\label{sec:SVGP}
Many schemes have been proposed to make inference using GPs scalable \cite{Snelson,Hensman,Hartikainen,Shen,Cunningham,Naish,Snelson2}. We use the stochastic variational inference (SVI) approach outlined in \cite{Hensman2} which uses an inducing point framework for variational approximation of the posterior $p(\mathbf{f}|\mathbf{y}, \mathbf{X})$ and marginal likelihood $p(\mathbf{y}|\mathbf{X})$ distributions. This method was chosen over alternatives because of its apparently high accuracy which can be balanced against computational complexity by tuning the number of inducing inputs. Note, all distributions below are implicitly conditioned on the observed inputs, $\mathbf{X}$, which we drop from the notation for brevity.
The approach begins at the inequality
\begin{equation}
\log p(\mathbf{y}|\mathbf{u}) \geqslant \mathbb{E}_{p(\mathbf{f}|\mathbf{u})}[\log p(\mathbf{y}|\mathbf{f})]
\label{eqn:inducing}
\end{equation}
where $\mathbf{u}=\{u_j\}_{j=1}^m$ are inducing variables: values of the output at the inducing input points $\mathbf{Z}=\{\mathbf{z}_j\}_{j=1}^m$. $m$ is the number of inducing inputs. Denoting the variational approximation to $p(.)$ as $q(.)$, the standard variational inequality is 
\begin{equation}
\log p(\mathbf{y}) \geqslant \mathbb{E}_{q(\mathbf{u})}[\log p(\mathbf{y}|\mathbf{u})] - \mathrm{KL}[q(\mathbf{u})||p(\mathbf{u})]\,\,.
\label{eqn:variational}
\end{equation}
Combining (\ref{eqn:inducing}) and (\ref{eqn:variational}), and exploiting the factorisation of the likelihood, $p(\mathbf{y}|\mathbf{f})=\prod_{i=1}^np(y_i|f_i)$, can be shown to give the bound
\begin{equation}
\log p(\mathbf{y}) \geqslant \sum_{i=1}^n \mathbb{E}_{q(f_i)}[\log p(y_i|f_i)] - \mathrm{KL}[q(\mathbf{u})||p(\mathbf{u})]\,\,.  \label{eqn:final_bound}
\end{equation}
As $f\sim\mathcal{GP}(\mu,k(.,.))$ identities for the derivatives of Gaussian-expectations (see \cite{Opper}) can be exploited to find a convenient form for the derivative of the bound. The derivatives with respect to the parameters of the variational distribution, inducing inputs $\mathbf{Z}$ and covariance hyperparameters can be found, and thus their values optimised by gradient descent. The predictive distribution can then be calculated by integrating the approximate posterior over $\mathbf{u}$ and $\mathbf{f}$ for a novel input $\mathbf{x}^*$.

\section{Related Work}
\label{sec:lit_review}
A large number of early studies focused on ambulance demand modelling investigated how quasi-static socio-economic attributes of a geographic region correlated with the ambulance demand using linear models \cite{Aldrich, Kvalseth,Cadigan,Schuman,Kamenetzky,Siler,Deems}. While these methods are of interest in understanding the causes of ambulance demand, they are not directly relevant to short-term demand prediction.
Nonparametric time-series methods have been implemented for forecasting in this context. A non-exhaustive list includes autoregressive moving average (ARIMA, \cite{Tandberg}); vector autoregression (VAR) to correlate demand at multiple sites \cite{Jones}; Singular Spectral Analysis (SSA, \cite{Vile}) and Winters' triple exponential smoothing \cite{Baker}. MEDIC is a common benchmark approach which is discussed in \cite{Setzler}. MEDIC exploits seasonality at different scales; the prediction for a given hour is the average of the call rates observed in the same hour of the previous weeks over the past month, and those same hours in previous years. MEDIC was said to be industry practice for Mecklenberg County, North Carolina \cite{Setzler}.
Kernel density estimates (KDEs) are used for demand prediction over continuous geographic space and hourly timebins in \cite{Zhou1} and \cite{Zhou2}. Predictions are generated by first creating a spatial field made up of  two-dimensional Gaussian kernels, one at each historic observation. They are then weighted for a given test input according to their geographic proximity, similarity in hour-of-day and similarity in hour-of-week. This method is extended for Melbourne in \cite{Zhou2}. Melbourne has a highly irregular city boundary which motivates \emph{kernel warping}: tweaking the kernel by projecting it onto the graph Laplacian of previous call-out locations.
By observing the variation in different strengths of seasonality between locations, a time-varying Gaussian mixture model (GMM) is proposed in \cite{Zhou3}. The weight of each Gaussian component is modulated by the phase of a prediction time within the weekly cycle.
Multiple studies have implemented Artificial Neural Networks (ANNs) for call-out prediction, most notably in Taipei \cite{Chen,Liao} and Mecklenberg County, North Carolina \cite{Setzler}. While the studies for Taipei show promising accuracy at coarse test resolution, they struggle due to sparsity at the scale of resolution required here. The resolutions tested in \cite{Setzler} are closer to that required here. Using spatial grids with $4\times4$mile cells, the ANN is found to have slightly improved mean square error scores relative to MEDIC.
Given the above review, GPR is thought to be an appropriate method for the following reasons:
\begin{itemize}
    \item Bayesian inference ensures robustness in a poor signal-to-noise-ratio regime.
    \item The seasonality in demand observed in other studies can be captured with periodic covariance functions. Further, composing the overall covariance as the product of those for the temporal and spatial components allows location-specific seasonal patterns to be captured. 
    \item GPR uses principled model fitting, rather than direct averaging over historic call-out locations and times. This limits the effect of noise in the training data on predictions.
    \item Standard GPR also predicts a distribution over call-densities, without requiring empirical approximation of uncertainty.
\end{itemize}

\section{Evaluation}
We use two methods to evaluate the predictive performance of the implemented models. Our predictions, which are estimates of the inhomogeneous Poisson rate, are denoted as $\lambda(\mathbf{s}, t)$ where $\mathbf{s}$ is the geographic space and $t$ the call time. The set of observed call times and locations is denoted $\mathcal{S}$.
\begin{figure*}
\centering
\subfig{0.49}{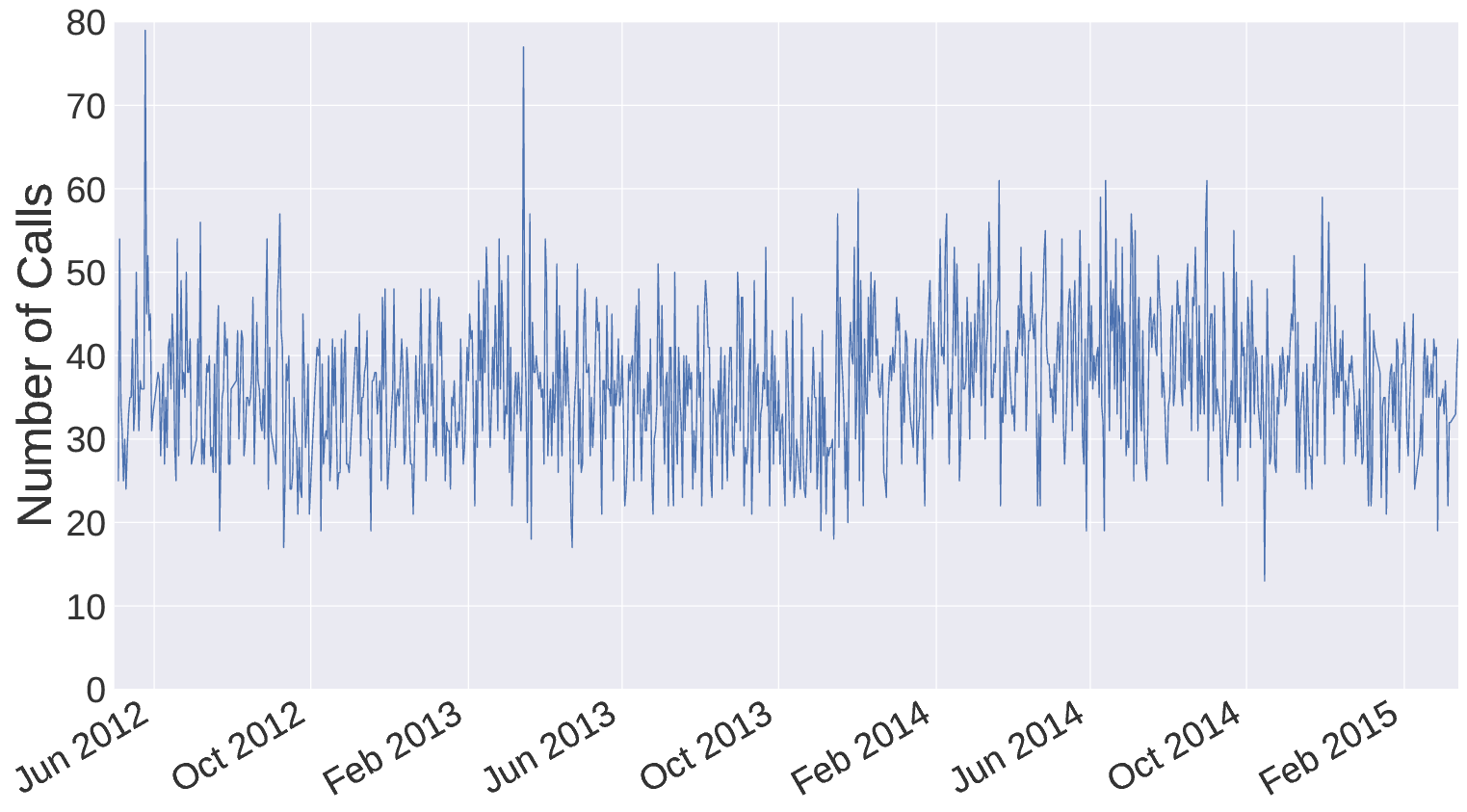}{The number of calls recorded daily over the training set.}{fig:daily_ts}
\subfig{0.485}{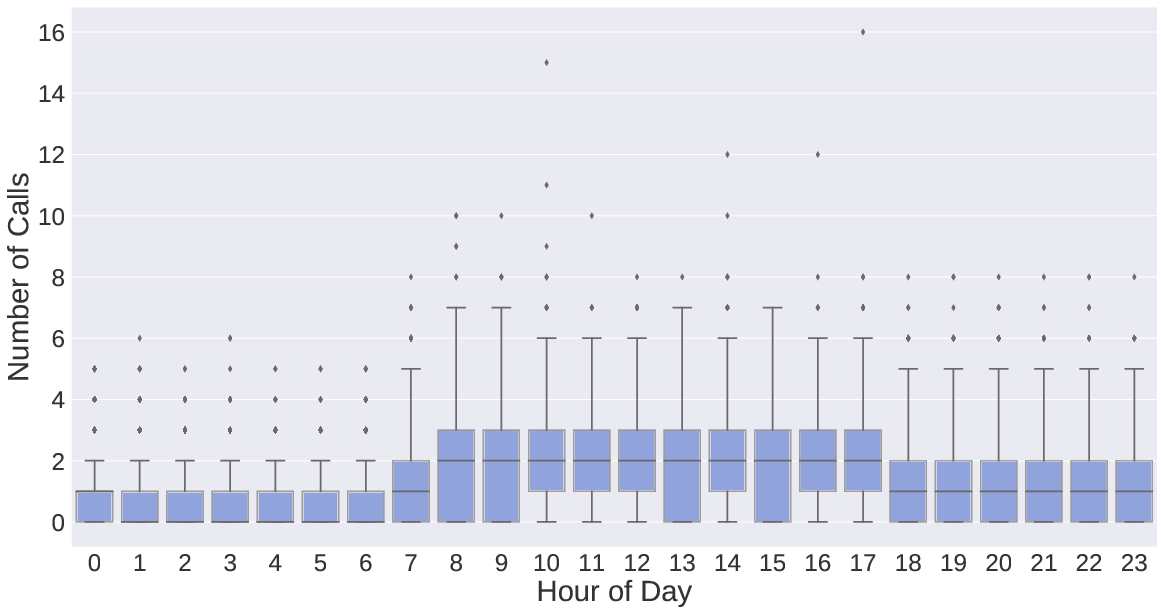}{Diurnal seasonality of ambulance demand in the training set.}{fig:hour_of_day}
\caption{The temporal variation of call-out rates.}
\label{fig:call_times}
\end{figure*}
\begin{enumerate}
\item {\bf Inhomogeneous Poisson Log-likelihood Density.}  This measure reflects the probability of a set of call outs, given the predictions $\lambda(\mathbf{s},t)$ are the true underlying rate of the Poisson process. Its mathematical form can be found by taking the product of probabilities of zero events over intervals with no events and probabilities of single events over small intervals $\delta\mathbf{s}\delta t$ around call-out times and locations. Taking the limit $\delta\rightarrow0$ and noting that only a single ordering of events is realised, the following likelihood density is found:
\begin{align}
l(\lambda(\mathbf{s},t)|\mathcal{S}) = \sum_{i=1}^k\log\lambda(\mathbf{s}_i,t_i) - \int_{\mathbf{s}}\int_t\lambda(\mathbf{s},t)\diff\mathbf{s}\diff t - \log k!
\label{eqn:ll}
\end{align}
Intuitively, we can interpret the first term as rewarding the prediction of high densities at event locations, the second term as penalising the overall prediction density and the final term as a normalising constant which reflects event ordering.
We approximate the integral as a summation over a 50$\times$50 spatial grid (covering the region shown in Fig. \ref{fig:call_locations_ct}) at 4-hourly resolution and evaluate the $\log k!$ term using the relation $\Gamma(k+1) = k!$ and the log-gamma function in Python programming language's SciPy package \cite{Scipy}. In general, STGP generates small but non-zero predictions in regions with low call density. However, MEDIC averages previous observations,  and will often predict zero events in a low density region. To avoid an undefined $\log0$ when an event occurs in a region with zero predicted counts, we clip our predictions $\lambda(\mathbf{s}_i,t_i)$ at $10^{-4}\mathrm{km}^{-2}\mathrm{hour}^{-1}$ when evaluating the first term in (\ref{eqn:ll}).
\item {\bf Mean Absolute Error.} We also use a more easily interpretable measure of accuracy. The mean absolute error (MAE) is given by
\begin{equation}
\mathrm{MAE} = \frac{1}{n}\sum_{i=1}^n\left|\lambda(\mathbf{s}_{i}, t_{i})A\tau - c_i\right|
\end{equation}
where $c_i$ is the count of events in bin $i$, with centre ($\mathbf{s}_{i}$, $t_{i}$); $A$ is the geographic area covered by a test bin and $\tau$ is the test timebin length. The $A\tau$ term rescales the predicted rate to a prediction of the number of counts within the test bin.
\end{enumerate}
While MAE is straightforward and readily communicable to a non-technical audience, its value is dependent on how the data are binned. This introduces subjectivity into the measurement. Conversely, the log-likelihood is independent of how the input space is descritised.
As it is assumed that ambulances will not be called-out to locations at sea, so any bins which do not cover any land are removed from the MAE calculation. Similarly, the geographic element of the integral $\int_{\mathbf{s}}\int_t\lambda(\mathbf{s},t)\diff\mathbf{s}\diff t$ is calculated over land only.

\section{Materials}
The dataset comprises around $4.5\times10^4$ call-outs over the Western Cape of South Africa between 4th January 2012 and 9th December 2015. However, coverage is sporadic in the first and last few months. The coverage is constant between May 2012 and 15th September 2015, and so any data outside this interval are removed.
The daily call rates are relatively stable over the training set (Fig. \ref{fig:daily_ts}), with the exception of a couple of spikes. There is no clear annual seasonality. On average, there are around 40 calls per day. More interesting is the hour-of-day seasonality, shown in Fig. \ref{fig:hour_of_day}. There are two observations here. First, there is significant hour-of-day periodicity, with higher demand close to and during working hours (around 8am to 5pm) and low demand overnight. Second, is that the distributions of call rates are positively skewed, consistent with the expected Poisson distribution.
 
\begin{figure*}
\centering
\subfig{0.54}{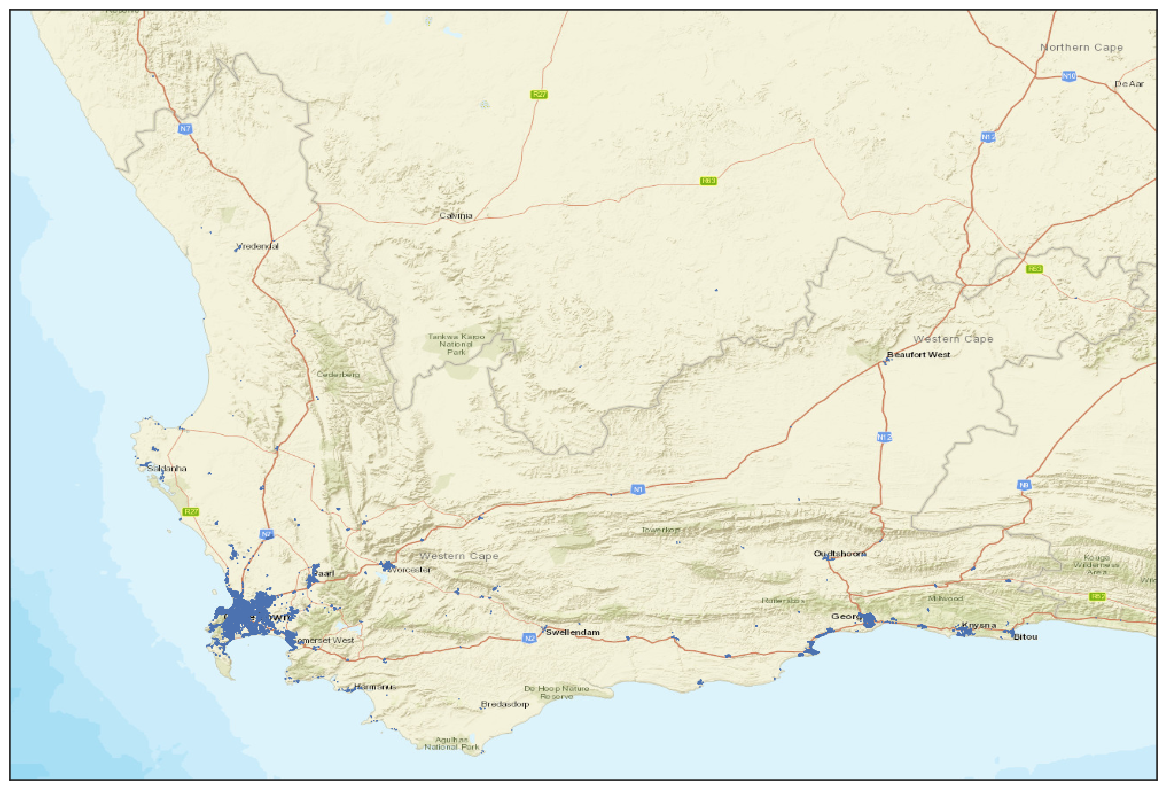}{Western Cape}{fig:call_locations_wc}
\subfig{0.43}{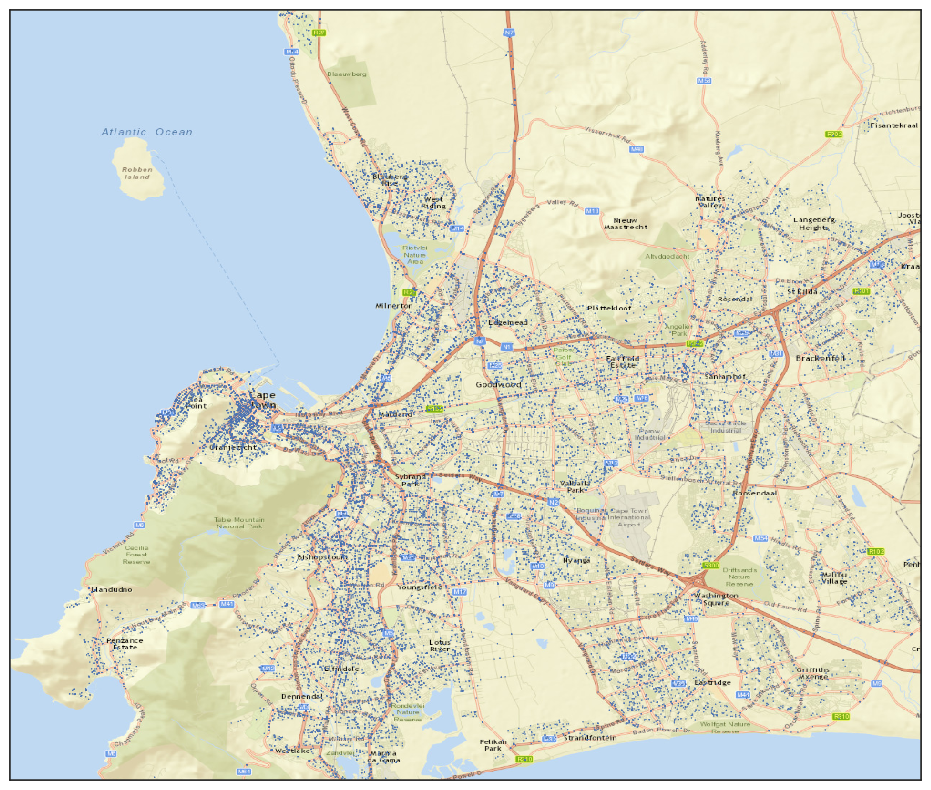}{Cape Town}{fig:call_locations_ct}
\caption{The locations of call-outs in the training set.}
\label{fig:call_locations}
\end{figure*}

The locations of all calls within the training set are presented in Fig.~\ref{fig:call_locations}. They are heavily concentrated in Cape Town relative to other areas of the Western Cape, in most of which there are almost no call-outs. Demand within Cape Town is focused around the downtown area north of table mountain.

\section{Method}
Here we present details of the kernel function and algorithm we use in our implementation of SVI LGCP which make up STGP. All analysis was carried out in the Python programming language, and the GPy library \cite{gpy} was used for Gaussian process implementation\footnote{We used \href{https://www.python.org/downloads/release/python-2712/}{Python 2.7.12} in conjunction with \href{https://github.com/SheffieldML/GPy/releases/tag/v1.8.4}{GPy 1.8.4.}}.

\subsection{Preprocessing}
\label{sec:preprocessing}
Given the concentration of demand in Cape Town, we restrict the geographic domain to its approximate limits\footnote{We define the limits of Cape Town to be a rectangle NSEW aligned, with a lower  left-hand corner at latitude -34.98, longitude 17.09 and upper right-hand corner of -30.16, 24.27 (the area covered by Fig.~\ref{fig:call_locations_ct}).}. As the demand is greatest in this region, we expect accurate predictions in the region to be most valuable.
The dataset contains a small number of non-emergency events such as inter-hospital transfers. As these are likely to obscure the true distribution of emergency events, they are removed.
Sparsity is a recognised problem in other studies \cite{Zhou1,Zhou2,Zhou3,Setzler}. Our dataset comprises an average of around 1.5 calls per hour over the whole Western Cape, around 1.1 calls per hour within Cape Town. This translates to average rates of around $1.2\times10^{-5}$ and $6.9\times10^{-4}$ km$^{-2}$hour$^{-1}$ respectively. For training we divide the geographic domain (see Fig.~\ref{fig:call_locations_ct}) into a 6$\times$6 grid within which events are counted for each 4 hour timebin. The dimensions of a cell in such a grid are 6.7km$\times$6.7km$\times$4hours. While this is a coarser level of descritisation than might be desirable, a finer grid was found to result in a poor signal to noise ratio which resulted in non-convergent training.

\subsection{Covariance Function}
Choice of covariance function is both a benefit and a drawback of using Gaussian process regression: it allows us to encode some high-level prior knowledge of the problem within our, but also introduces a seemingly arbitrary decision. While a method to remove subjectivity from this process have been proposed  \cite{Duvenaud}, it involves multiple training cycles which is not practical when the training set is large. Instead we use what is already known about the problem to inform our choice as far as possible.

\subsubsection{Temporal Covariance}
While the daily call rate seems not to exhibit any significant variation over the training set (Fig.~\ref{fig:daily_ts}), there is considerable hour-in-day seasonality (Fig.~\ref{fig:hour_of_day}). Inspection of day-of-week and month-of-year seasonality did not yield any notable observations. We thus choose a single periodic kernel \cite{MacKay} for the temporal component of the input:
\begin{equation}
k_t(t,t') = \theta_t \exp \left[  - \frac{1}{2l_t^2}
        \sin^2\Big(\frac{\pi (t - t')}{T} \Big) \right] 
\end{equation}
where $l_t$ is the lengthscale of the process; $\theta_t$ reflects its variance and $T$ is the period of oscillation, in our case 24 hours.

\subsubsection{Spatial Covariance}
The spatial distribution of events (Fig.~\ref{fig:call_locations_ct}) shows there is significant variation over the domain and some challenges caused by boundaries in the landscape. We choose the commonly used radial basis function (RBF) kernel for its simplicity and robustness:
\begin{equation}
k_s({\bf s}, {\bf s}') = \sigma_s^2 \exp\left(-\frac{\sum_j^m(s_j - s_j')^2}{l_s^2}\right)
\end{equation}
where $\sigma_s^2$ is the variance, $m$ is the number of dimensions (two in our case) and $l_s$ is the spatial lengthscale. We choose a relatively small value of $l_s$ in the hope that the issues caused by geographic boundaries may be mitigated. 

\subsubsection{Hyperparameter Tuning}
Given the sparsity in the data, and therefore the poor signal to noise ratio, we expect the surface of the marginal likelihood, $p(\mathbf{y}|\mathbf{x})$ to be highly non-convex. This makes hyperparameter tuning by gradient descent methods problematic. We thus fix covariance hyperparameters at values known to be sensible from exploratory analysis and prior understanding, for all except the variance. These values are given in Table~\ref{tab:hyperparam_values}.
\begin{table}
\centering
\begin{tabular}{rcl}
\toprule
{\bf Name} & {\bf Symbol} & {\bf Value}\\\midrule
Time Period & $T$& 24 hours\\
Time Lengthscale & $l_t$& 8 hours\\
Spatial Lengthscale & $l_s$ & 10 km\\\bottomrule
\end{tabular}
\caption{Fixed covariance hyperparameter values.}
\label{tab:hyperparam_values}
\end{table}

\begin{figure*}
\centering
\subfig{0.45}{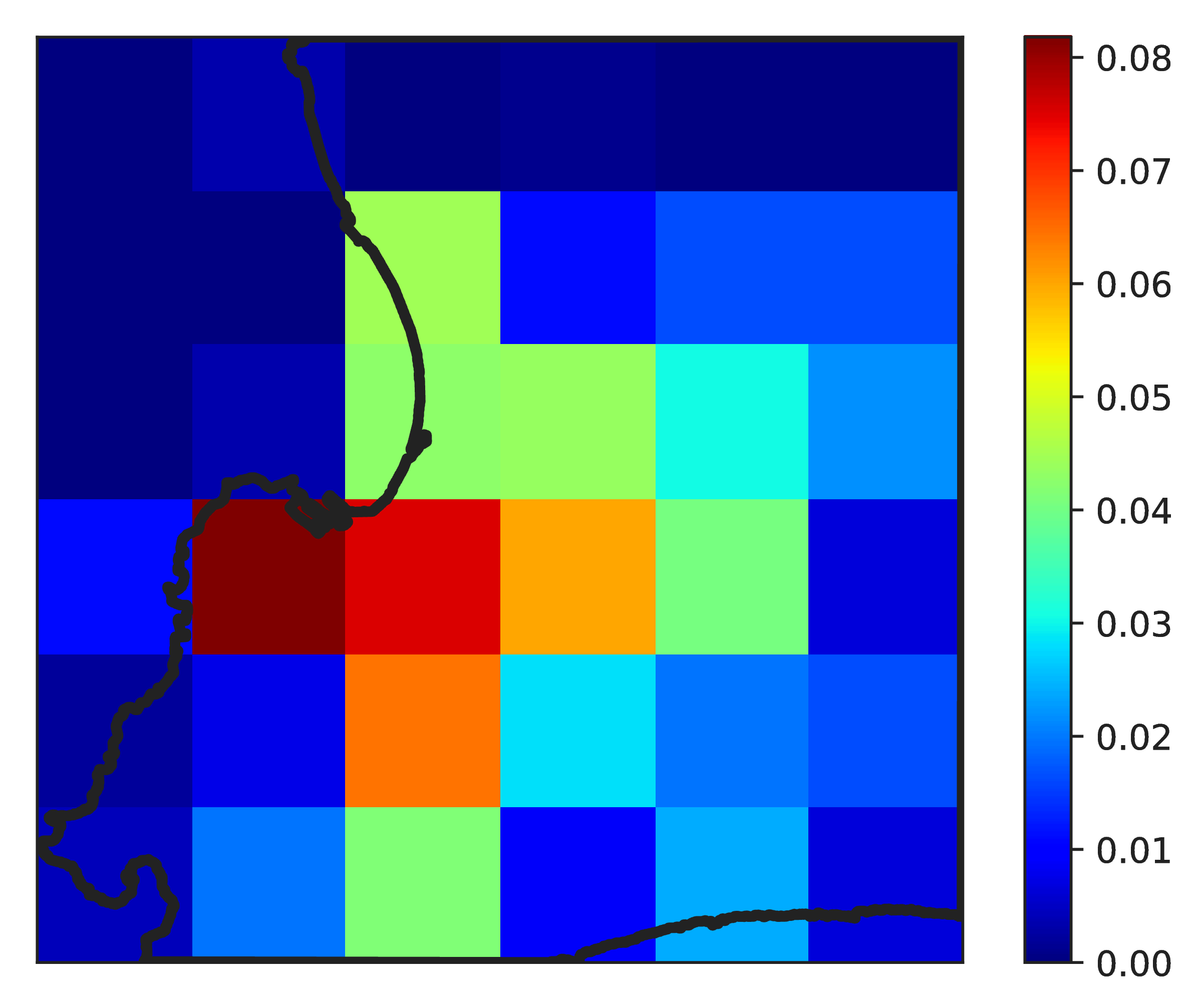}{0-1am.}{fig:hour_0}
\subfig{0.45}{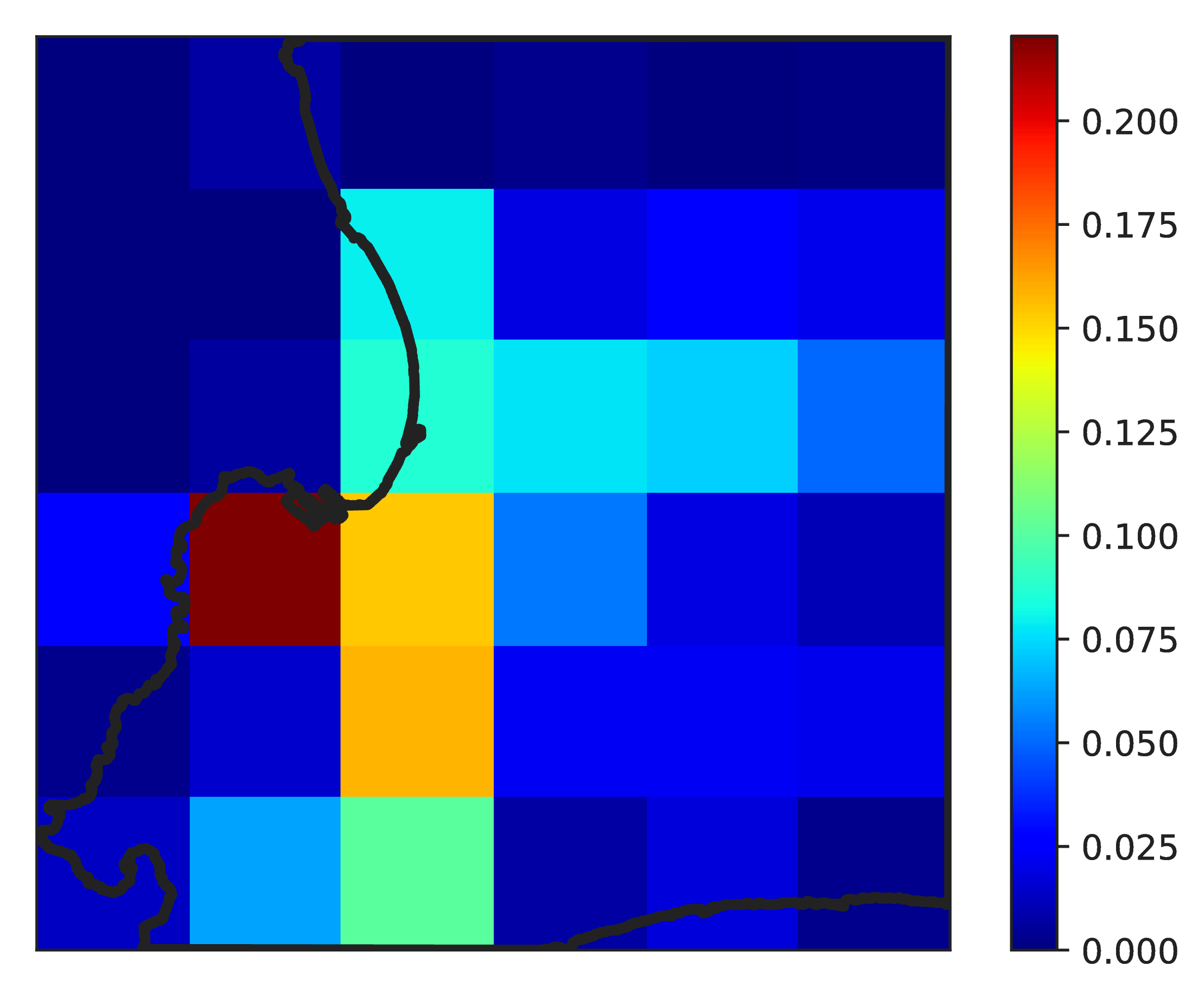}{3-4pm.}{fig:hour_15}
\caption{The empirical distributions of spatial demand for the hours of day as labelled, averaged over all days in the training set.}
\label{fig:empirical_spatial}
\end{figure*}

\subsubsection{Composition}
As described in \cite{Duvenaud}, the sum or product of two valid kernels is itself a valid kernel, however summation and multiplication give rise to different behaviours. Multiplication captures interactions between kernels whereas summation models each as an independent process. This is analogous to AND and OR operators enacted on the similarity measured by each kernel \cite{Duvenaud}. Deciding which is appropriate is dependent on the level of spatiotemporal interaction in the data.
Fig.~\ref{fig:empirical_spatial} shows the spatial distribution of calls in the training set. Fig.~\ref{fig:hour_0} shows that between midnight and 1am, while Fig.~\ref{fig:hour_15} shows that between 3 and 4pm, and both are averaged over all days in the training set. While the general shape of the distribution is similar in both cases, there is a greater focus of event density in the downtown region between 3 and 4pm. We thus conclude that there is some spatiotemporal interaction.
We choose the following kernel structure:

\begin{equation}
    k_{s,t}(\mathbf{x},\mathbf{x}') = k_{t, m}(t,t') + k_{s, m}(\mathbf{s},\mathbf{s}') + k_{t, i}(t,t') \times k_{s, i}(\mathbf{s},\mathbf{s}')
\end{equation}

where $\mathbf{x}=(\mathbf{s},t)$, $\mathbf{x}'=(\mathbf{s}',t')$, both $k_{t, m}(.,.)$ and $k_{t, i}(.,.)$ are instances of $k_t(.,.)$ without tied parameters, the same is true for $k_{\mathbf{s}, m}(.,.)$ and $k_{\mathbf{s}, i}(.,.)$. This composition is motivated by the majority of variance being marginal in either time or space, which can be captured by the first two components. We hope to capture some of the residual variance arising from spatiotemporal interaction using the product kernel.

\subsection{Inducing Inputs}
While the inducing point framework provides an interpretable trade-off between runtime and accuracy, it requires selection of the number of inducing inputs and their initial values. With SVI GPs \cite{Hensman2}, it is common practice to optimise the inducing inputs by gradient descent of the marginal likelihood surface. However, we find our marginal likelihood surface to be largely non-convex and optimisation in this manner to result at poor values of inducing inputs. To mitigate this, we randomly sample, without replacement, 180 inputs from the training set for use as inducing inputs. We find this number of points sufficient to achieve a good model fit. 
 
\subsection{Gradient Descent}
As discussed in Section~\ref{sec:SVGP}, we train STGP using a variant of gradient descent. There are a number of possible algorithms which could be used. We use the Broyden-Fletcher-Goldfarb-Shanno (BFGS) algorithm \cite{Fletcher} as implemented in in Python language's SciPy library \cite{Scipy}. We find this converges in less time than Newton's method and scaled conjugate gradients.
\begin{figure*}[h]
\centering
\subfig{0.45}{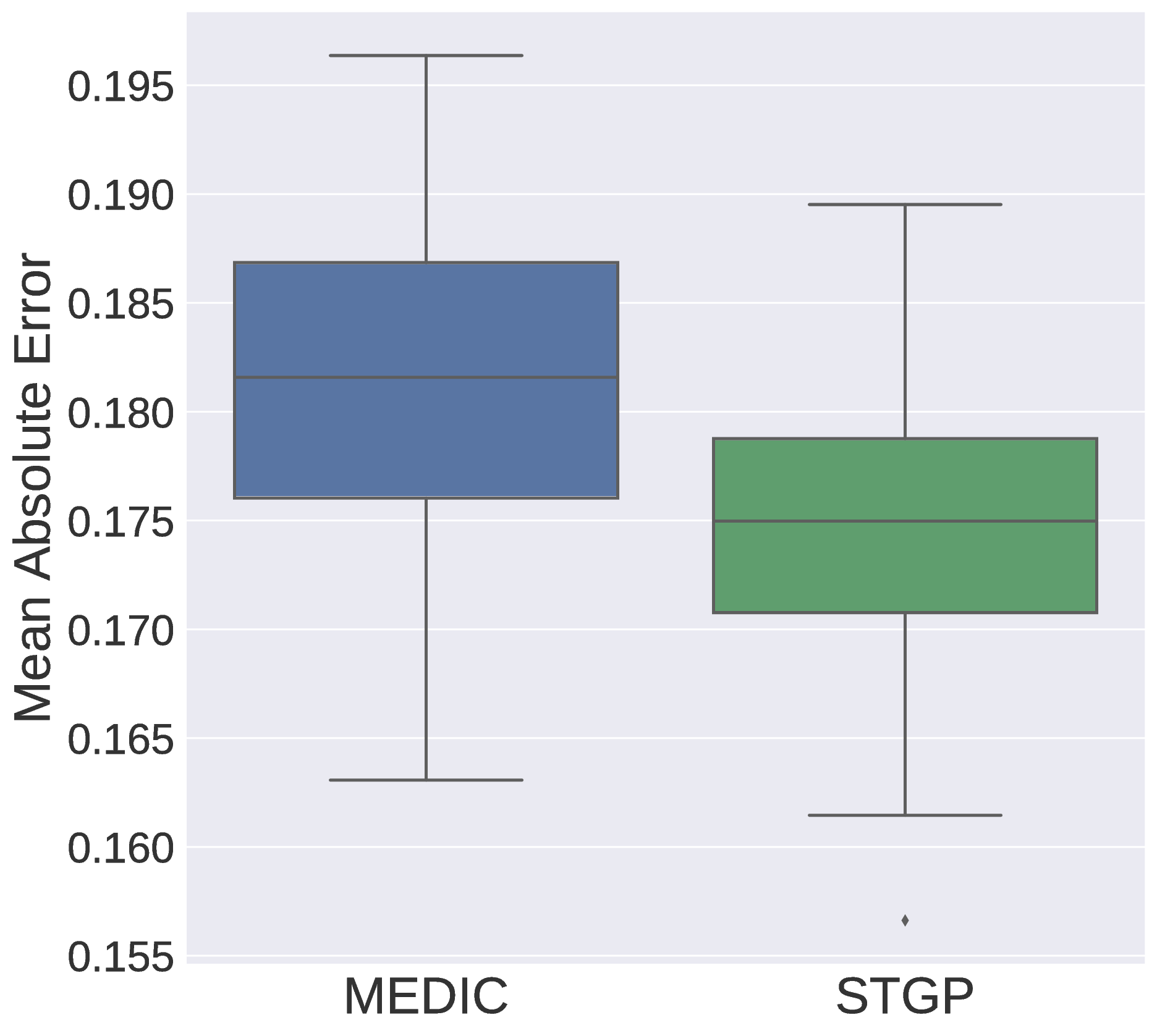}{}{fig:mae}
\subfig{0.45}{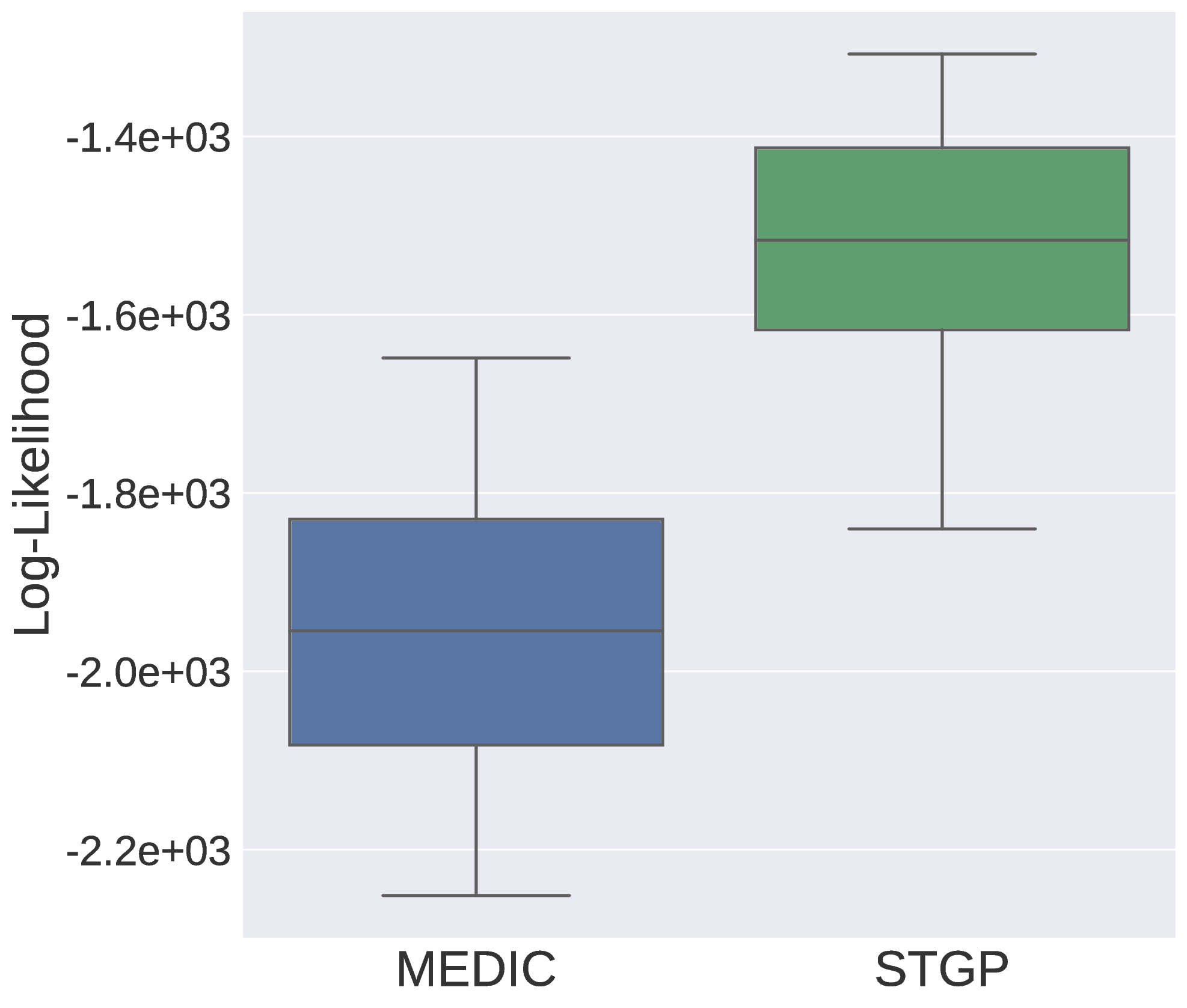}{}{fig:ll}
\caption{The distribution of weekly predictive accuracies.}
\label{fig:accuracy}
\end{figure*}

\subsection{Benchmark}
As discussed in Section~\ref{sec:lit_review}, MEDIC is a benchmark of call-out prediction accuracy which has been applied in industry and is used in \cite{Zhou1,Zhou2,Setzler}. The method predicts the counts  to be the average of those for the same timebin in the previous four weeks over the past 5 years \cite{Setzler}. This is generalised to the spatiotemporal domain, by applying it within each geographic area. However, no consensus could be found as to the size of each bin, which appears to be context dependent. \cite{Setzler} implements both 2$\times$2 mile and 4$\times$4 mile descritisation as both 1 and 3 hour intervals. \cite{Zhou1} and \cite{Zhou2} use 1 km$^2$ and 1 hour bins. Here, we divide the spatial domain into a 6$\times$6 grid and implement MEDIC within each, over the range of training data available ($\sim$3 years). We use this spatial resolution in conjunction with a 4-hourly temporal resolution as a compromise between capturing the spatiotemporal dynamics, and reducing the affect of noise through averaging.

\subsection{Algorithm}
To generate predictions over the test set (17th March to 14th September 2015 inclusive), we use a weekly rolling train/test scheme as described in Algorithm~\ref{algo:rolling}. The training set size for STGP is clipped at 6 months to reduce runtime. To ensure against a poor STGP model fit resulting from gradient descent converging to a local minimum, we assess the model likelihood after training. If the training log-likelihood falls below a set threshold ($-1.07\times10^{4}$), we retrain the model with a different random seed.

\begin{algorithm}
  \caption{Rolling train/test scheme}
  \label{algo:rolling}
  \begin{algorithmic}[1]
      \State $S \gets \mathrm{load\_call\_times\_locations()}$
      \State $D \gets \mathrm{spatiotemporal\_binning}(S)$
      \State $t_{te} \gets$ 00:00 17th March 2015
      \State $t_{end} \gets$ 23:59 14th September 2015
      \State $P \gets [\,]$
      \While {$t_{te} < t_{end}$}
       \State $D_{tr} \gets D[\mathrm{where}( t < t_{te})]$
       \State $D_{te} \gets D[\mathrm{where}( t_{te} \leqslant t < t_{te} + \mathrm{1 week})]$
       \State $M \gets \mathrm{get\_trained\_model}(D_{tr})$
       \State $P_{te} \gets \mathrm{get\_predictions}(M,D_{te})$
       \State $P \gets \mathrm{append}(P,P_{te})$
       \State $t_{te} \gets \min(t_{te} + \mathrm{1 week},\,\,t_{end})$
       \EndWhile
  \end{algorithmic}
\end{algorithm} 

\section{Results}
The test-set accuracies are shown in Table~\ref{tab:overall_acc} and the distributions of weekly accuracies are given in Fig.~\ref{fig:accuracy}. STGP exhibits superior predictive power according to both MAE and log-likelihood, for all quantiles. The difference in performance is most marked when measured by log-likelihood. This is likely to be caused by two differences between STGP and MEDIC. First, STGP predicts a smooth spatial distribution at each timestep, whereas MEDIC is restricted to the spatial grid used for training. This allows STGP to predict in a more accurate manner, with less predictive weight being wasted in in locations with few call-outs adjacent to locations of high density. Such an effect would cause STGP to have a smaller integral of predicted rate over time and space, and/or greater predicted rates at probable call locations. Further, events within bins where MEDIC has predicted a zero rate still reduce its test set log-likelihood score. While we round such predictions up to $10^{-4} \mathrm{km}^{-2} \mathrm{hour}^{-1}$, they still create significantly negative contributions to the summation $\sum_{i=1}^n\log\lambda(\mathbf{s}_i, t_i)$.

\begin{table}[b]
\centering
\begin{tabular}{rcl}
\toprule
{\bf Method} & {\bf MAE} & {\bf Log-likelihood}\\\midrule
STGP & $0.174$ & $-6.83\times10^4$\\
MEDIC & $0.181$ & $-7.96\times10^4$\\\bottomrule
\end{tabular}
\caption{Predictive accuracies over the full test set.}
\label{tab:overall_acc}
\end{table}

\begin{figure*}[h]
    \centering 
    \includegraphics[width=0.9\textwidth]{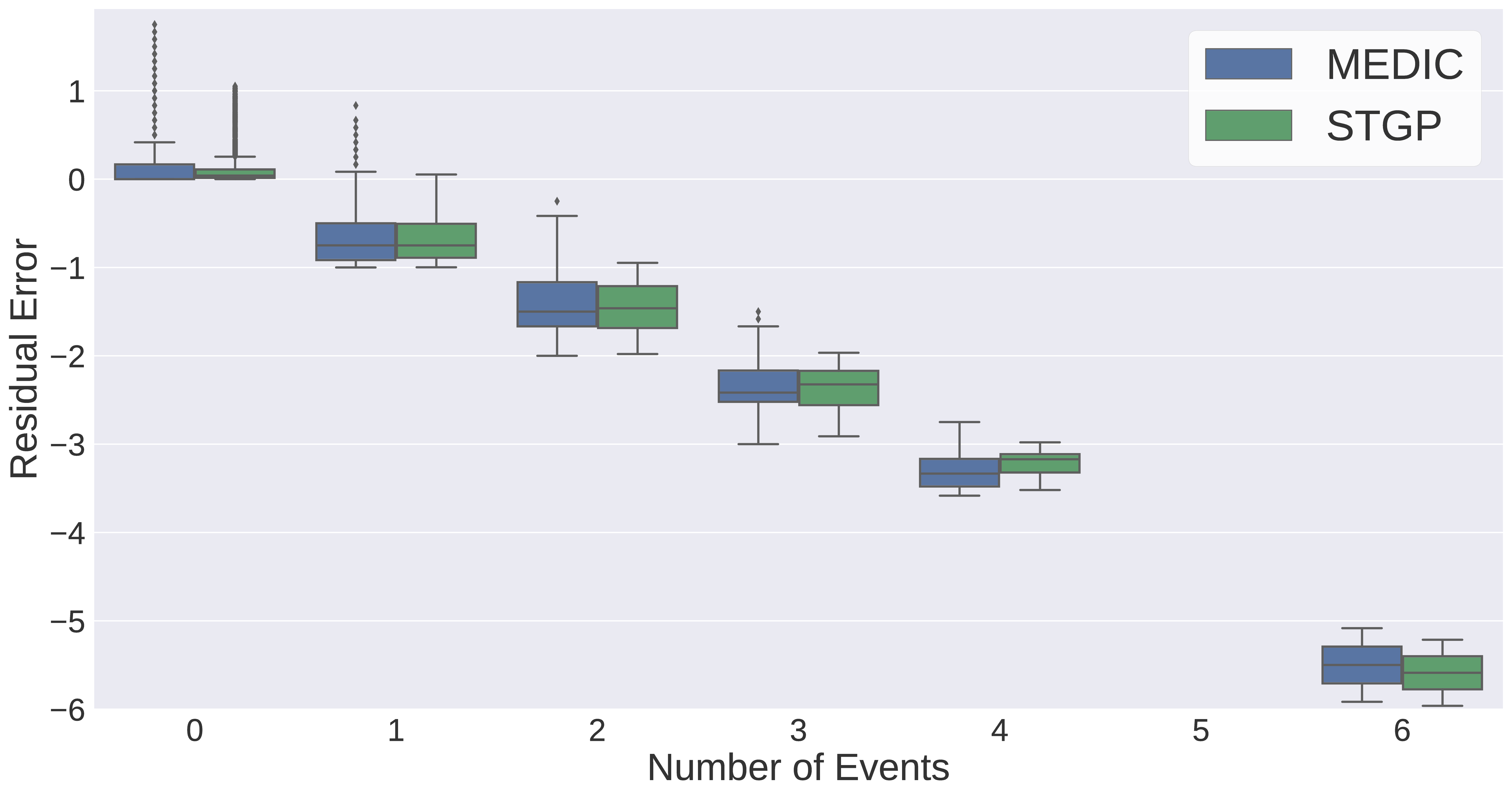}
    \caption{The distribution of residuals for each value of events per bin.}
    \label{fig:residual_vs_num_events}
\end{figure*}

\begin{figure*}
\centering
\subfig{0.45}{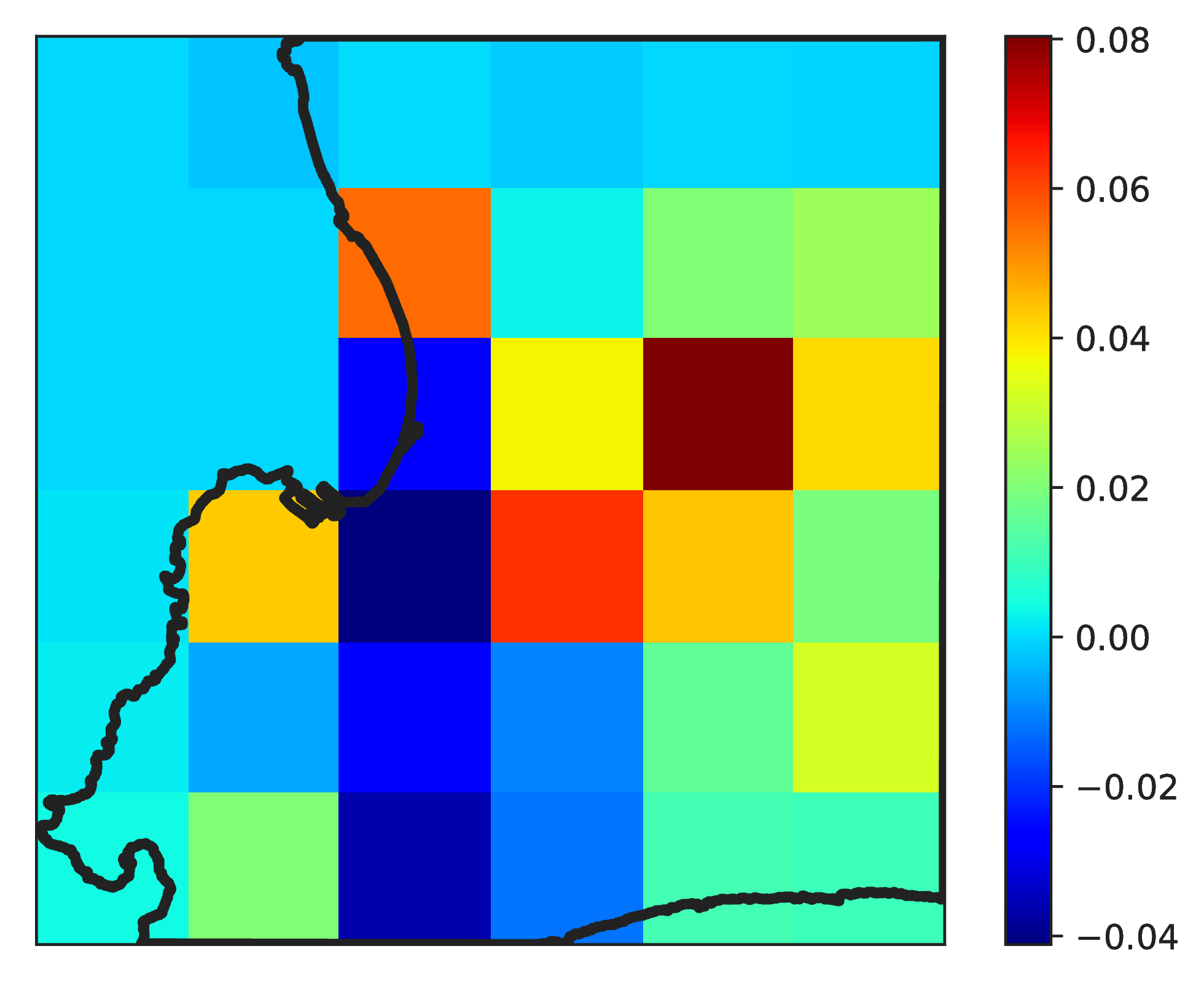}{MEDIC}{fig:spatial_residual_medic}
\subfig{0.45}{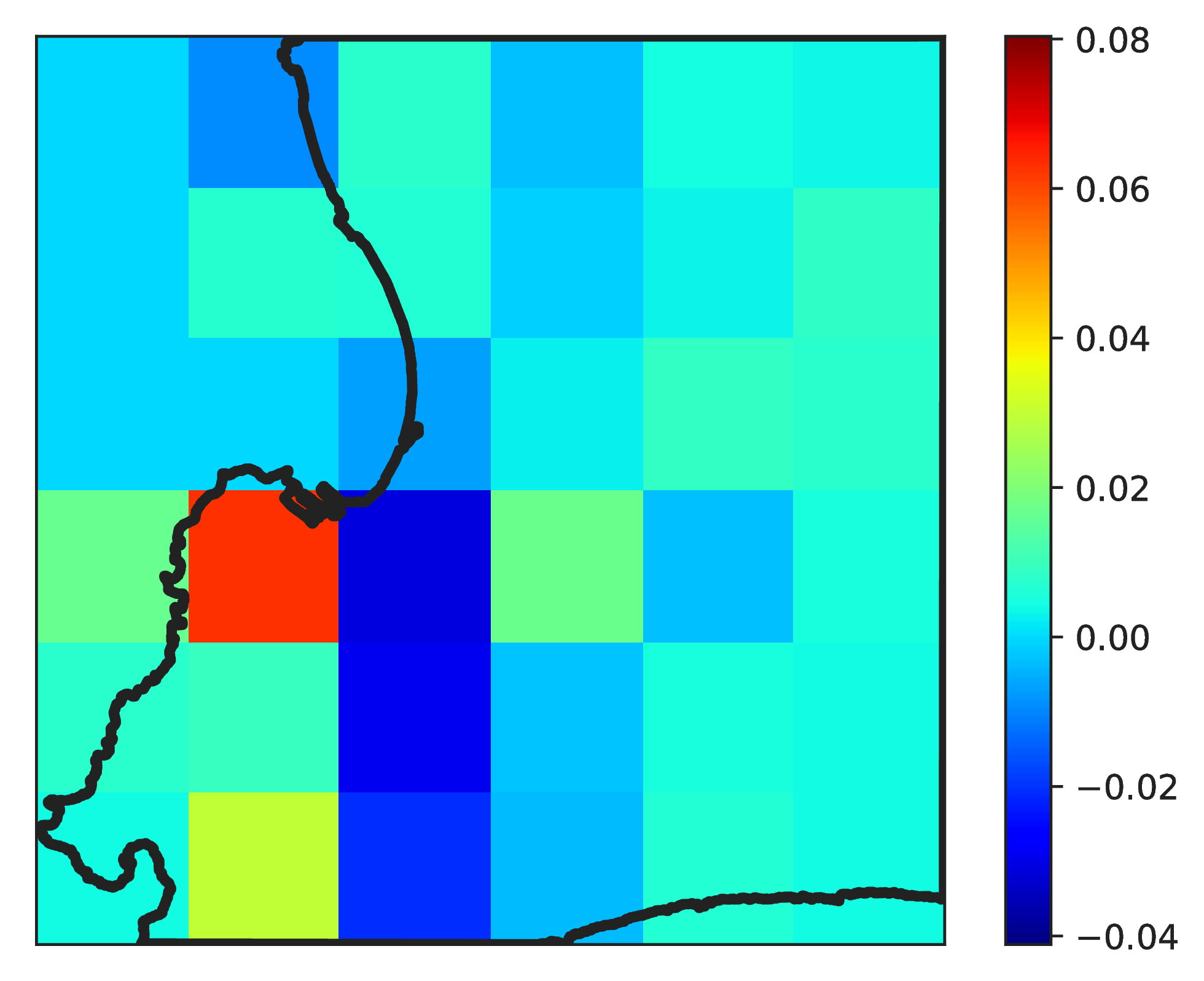}{STGP}{fig:spatial_residual_stgp}
\caption{The spatial distribution of average residuals.}
\label{fig:spatial_residuals}
\end{figure*}

Exploring further the dependence of the accuracy on the value of the output , the distribution of residuals for all values of events per bin (during the MAE calculation) is given in Fig.~\ref{fig:residual_vs_num_events}. There are a number of comments to make here. First, STGP is much more accurate when it comes to predicting for bins with no events. It is likely that MEDIC is less accurate in these cases due to its quantised predictions: it predicts averages of relatively few noisy count observations, so even if one of the historic observations is non-zero, the resulting inaccuracy is significant. Average accuracies for bins containing events are more similar, however STGP is more consistent, with less variance in accuracy. It should be noted that neither method is accurate for bins with large number of events. However, Fig.~\ref{fig:residual_vs_num_events} is misleading as it does not give an indication of the relative prevalence of each number of events. The large majority of bins contain zero events, few contain a single event and very few contain more than one event. It is expected that such skew results in under-prediction for bins with many events.   
Continuity of the input and output result in smaller bias in STGP predictions, relative to MEDIC. The average spatial distribution of residuals over the test set is presented in Fig.~\ref{fig:spatial_residuals}. In general, MEDIC exhibits positive forecast bias in areas of low demand. This is thought to result not only from quantisation of the output, but no encoding for spatial similarity between adjacent bins. The spatial kernel used in STGP enforces smoothness which reduces the effect of noise. Note also that MEDIC shows greater negative bias in one of the downtown regions with greatest call density. This is concerning as it could result in under-provision of resources to high-demand areas.

\section{Conclusion}
We have presented STGP, a novel method for spatiotemporal ambulance demand using Gaussian process regression (GPR). GPR has a number of desirable features in this context such as the ability to choose classes of pattern which the model can fit, and make predictions with uncertainty. Our method shows promise, significantly outperforming MEDIC, a commonly used benchmark, by both MAE and log-likelihood. In addition, STGP is more consistent, with less variability in accuracy.
This work could be extended in a number of directions. Exogenous datasets relating to factors which influence ambulance demand (historic road traffic accidents, population density etc) may be used to increase prediction accuracy within a multi-task GPR framework (such as that described by \cite{Bonilla}). The geography in Cape Town is challenging due to an irregular coastline and mountainous park areas which are largely uninhabited, causing a jagged spatial distribution of demand. This issue is addressed for Melbourne in \cite{Zhou2} by using kernel-warping - projecting the predicted spatial distribution onto the distribution of sampled historic call outs. This could be augmented with the road networks and/or samples from the population density distribution. Such kernel warping would likely increase the accuracy of STGP.
Methods to exploit the predictions might also be considered. For example, the predictive distribution, marginalised over space, is a forecast for total demand at that timestep. Percentiles of this distribution could be translated into possible staffing levels, with different levels of provision for different levels of certainty. Conversely, the predictions could be used as an indication of environmental dynamics in a planning problem, in which the resting places of ambulances between calls are optimised at different times of day.

\section*{Acknowledgements}
The authors would like to thank \href{https://www.er24.co.za/}{ER24} (South Africa's largest private emergency services operator) and \href{http://www.mediclinic.co.za/}{Mediclinic}, of which ER24 is a wholly owned subsidiary, for providing the data used in the study to InformedActions and Grove Group, as well as their collaboration and contribution to this research. They would also like to thank \href{https://www.groveis.com/}{Grove Group}, for their financial support and assistance with this publication and the practical application of the work it describes. None of InformedActions, Grove Group, ER24 or Mediclinic influenced the analysis or interpretation of data; the writing of the manuscript; or the decision to submit the manuscript for publication.

\section*{Conflict of Interest Statement}
Seth Nabarro and Tristan Fletcher are shareholders/employees of InformedActions, which is a consultant to ER24.

\bibliographystyle{plainnat}

\end{document}